\documentclass[conference]{IEEEtran}

\usepackage{cite}
\usepackage{amsmath,amssymb,amsfonts}
\usepackage{algorithmic}
\usepackage{graphicx}
\usepackage{multirow}
\usepackage{subcaption}
\usepackage{booktabs}
\usepackage{textcomp}
\usepackage{xcolor}
\def\BibTeX{{\rm B\kern-.05em{\sc i\kern-.025em b}\kern-.08em
    T\kern-.1667em\lower.7ex\hbox{E}\kern-.125emX}}
\begin{document}

\title{Using Genetic Algorithms to Simulate Evolution\\
}
\author{\IEEEauthorblockN{Manasa Josyula}
\IEEEauthorblockA{\textit{Sharon High School} \\
manasaj5a@gmail.com}\\

}

\maketitle
\begin{abstract}
    Evolution is the theory that plants and animals today have come from kinds that have existed in the past. Scientists such as Charles Darwin and Alfred Wallace dedicate their life to observe how species interact with their environment, grow, and change. We are able to predict future changes as well as simulate the process using genetic algorithms. Genetic Algorithms give us the opportunity to present multiple variables and parameters to an environment and change values to simulate different situations. By optimizing genetic algorithms to hold entities in an environment, we are able to assign varying characteristics such as speed, size, and cloning probability, to the entities to simulate real natural selection and evolution in a shorter period of time. Learning about how species grow and evolve allows us to find ways to improve technology, help animals going extinct to survive, and figure* out how diseases spread and possible ways of making an environment uninhabitable for them. Using data from an environment including genetic algorithms and parameters of speed, size, and cloning percentage, the ability to test several changes in the environment and observe how the species interacts within it appears. After testing different environments with a varied amount of food while keeping the number of starting population at 10 entities, it was found that an environment with a scarce amount of food was not sustainable for small and slow entities. All environments displayed an increase in speed, but the environments that were richer in food allowed for the entities to live for the entire duration of 50 generations, as well as allowed the population to grow significantly\cite{deb2002fast}.
\end{abstract}

\section{Introduction}
Genetic algorithms generate high quality solutions to search and optimization by utilizing biological ideas such as crossover, mutation, and natural selection. Designs or parameters are represented by species in an optimization problem which gives the genetic algorithms an objective to find the optimal design or parameters.  However, genetic algorithms are an excellent tool for directly studying evolutionary principles and behavior. We are able to put a species in different environments to watch the entities evolve and interact with the rules of the environment. Using genetic algorithms, we will be able to observe long term evolution of a species in a shorter amount of time, predict possible mutations that will allow the entities to evolve, and find ways to prevent a species from going extinct. Learning about this is important to keep species from endangerment and learning ways that they can mutate in the future to help them survive. Genetic algorithms can also be used for disease prevention by learning what environments certain bacteria can and can’t reproduce in. 
We implemented genetic algorithms to simulate evolution by first creating an environment which has entities. Each entity has a varied speed, size, and cloning probability. Additional points act as food, which allows us to assign a survival equation to the entities. The amount of food the entities collect each run determines whether they live for the next run, and if they would be able to clone themselves. Changing the amount of ‘food’ in the environment gives the entities a different situation to try surviving in. We can collect data from the entities such as changes in speed, size, cloning, and population and display the data on graphs so that we can see what the entities are doing to survive. After testing the environment with 100, 200, and 300 starting food counts, we found that the 300 starting food count is the best to sustain a species over 50 generations because 100 and 200 food to start was not enough, and did not allow the population to live for over 15 generations. Starting with 300 food allowed the species to grow constantly, and each run showed that the population lived for the entire 50 generations. The cloning probability was also at its highest here at around an .80 chance that the entities would clone. 
We developed a tool using genetic algorithms to simulate evolution, natural selection, and population growth so that observation can be done on the changes and growth of a species in relation to its environment. By adding more parameters, different species in the same environment, and the possibility of crossover, it is made possible to imitate interactions happening in real life, and is a faster way to observe evolution and natural selection.

\section{Relevant Work} \label{review}
Genetic algorithms are used to find the optimal design or parameters for an optimization problem. Moreover, genetic algorithms can be used for studying behavior and principles of different species. One of the general principles of genetic algorithms is having sequential generations, which allow us to see how a species evolves over time, as well as how it interacts with its environment. Genetic algorithms also look at fitness evaluation at each generation, which looks at what the speed, size, cloning probability, and other parameters of a species are at the moment, and thinking about how they have differences than the previous generation. Survival of the fittest, or natural selection, is also a big part of observing evolution and how a species interacts with its environment. By giving each entity a certain fitness at each generation, this presents the possibility of death. If an entity does not meet the requirements to live to the next day, which in this case is based on the amount of food it collects, it will die and therefore decrease the population. On the other hand, if an entity collects enough food to clone, it will reproduce. Reproduction also introduces the possibility of mutation and crossover. Mutation happens when there is damage or a change in a gene. This causes the newly produced entity to have a difference that no other entity in its species will have. This is the beginning of evolution. Crossover occurs when two chromosomes, one from the mother and one from the father line up and some parts of them can be exchanged. Crossover helps the shuffling of genetic material. In every environment, there is a carrying capacity, which is the amount of entities the environment can sustain without dying. This is an important idea because the natural world makes sure there are enough supplies for every species living in it, and if there are too many of one kind, the environment and entire food chain will collapse. To make sure that no species oversteps its carrying capacity, every environment includes a limited amount of abiotic factors (such as oxygen and water) and biotic factors (food) which will determine the carrying capacity for that environment. 

\section{Methodology} \label{methodology}
Our model holds a population with different fitness abilities which live in an environment with scattered food, who will compete in collecting food to survive each day and clone. If an entity does not collect enough food to survive, it will not live in the next simulation. When an entity collects enough food, it will be able to clone, and the clone will have a possibility to mutate.. This replicates evolution by displaying survival of the fittest where the strongest entities who have a higher probability of surviving will live and evolve into the best possible species for their environment. In order to use genetic algorithms to simulate evolution and natural selection, we start with an environment with a set population of entities. Each entity has a unique size, speed, and cloning probability parameter which is randomly selected between a set of possible values. In the environment, there are randomly placed plots, which are used as food for the entities. The motion of entities is based on the equation: 

\begin{equation}
    movement = f(speed, size) 
\end{equation}

The movement will be affected by both the speed and size of the entity so that each one moves uniquely. Every time an entity gets within range of a piece of food, it collects it and it adds to one point on the entity’s score. To resolve a tie, the AI randomly awards food to one of the nearby entities. The survival rate of the entities is based off of the survival equation:\\

\begin{equation}
    \frac {(5+pop.size[i])*(5+pop.speed[i])*(1+pop.cloning[i])}{36}
\end{equation}

When an entity collects enough food to clone, its clone will start in the same position as its original, and they will have similar characteristics as well. Every clone has a probability for mutation which was set to a .50 chance. If the clone mutates, it can mutate its speed, size, or cloning probability. The speed of the clone depends on the speed of the original. The new speed of the clone will be calculated based on the original’s speed, and if it is higher than the maximum speed, then the clone’s speed will be set to the maximum speed. If the calculated speed is less than the minimum set speed, then the clone’s speed will be equal to the minimum speed. If the calculated clone’s size based on the parent entity is higher than the maximum size, then the clone’s size will be set to the maximum size, and if it is lower than the minimum size, then the clone’s size will be equal to the minimum size. Similarly, if the clone’s calculated cloning probability is greater than 1, it will be set to 1, and if it is lower than 0, then the cloning probability of the clone will be set to 0. Since this model displays asexual reproduction, crossover is not part of the parameters. 

\section{Results} \label{Results}
 We ran several batches of simulations, and for each simulation we changed the initial food count variable. After running the simulation, we collected data for the average size, speed, and cloning probability of the entities for each generation, along with the population size, over 50 generations to put into a line graph. 
 \begin{figure*}[!htb]
    \centering
    \includegraphics[scale=.4]{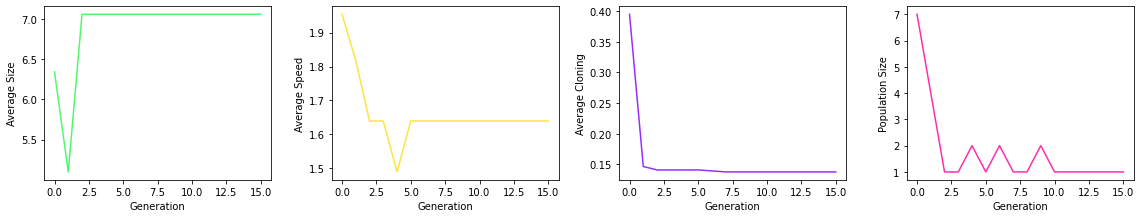}
    \caption{Population Average Speed, Size, Cloning Probability, and Population Growth With 100 Starting Food: Trial 1}
    \label{fig:100Food1}
\end{figure*}
\begin{figure*}[!htb]
    \centering
    \includegraphics[scale=.4]{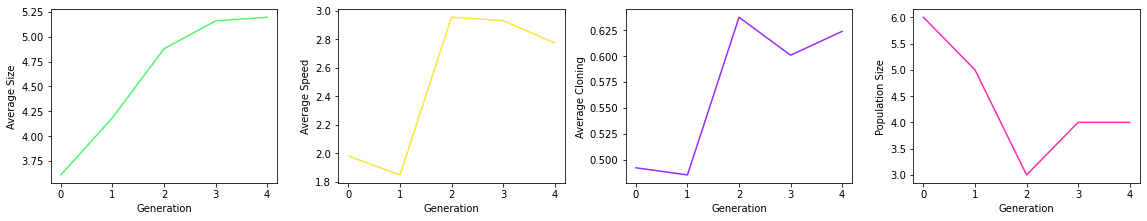}
    \caption{Population Average Speed, Size, Cloning Probability, and Population Growth With 100 Starting Food: Trial 2}
    \label{fig:100Food2}
\end{figure*}
\begin{figure*}[!htb]
    \centering
    \includegraphics[scale=.4]{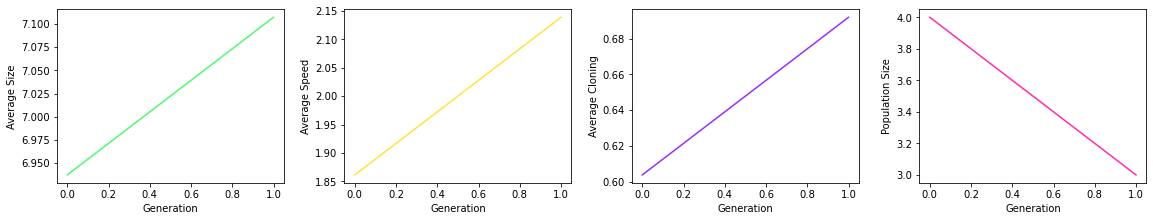}
    \caption{Population Average Speed, Size, Cloning Probability, and Population Growth With 100 Starting Food: Trial 3}
    \label{fig:100Food3}
\end{figure*}
\begin{figure*}[!htb]
    \centering
    \includegraphics[scale=.4]{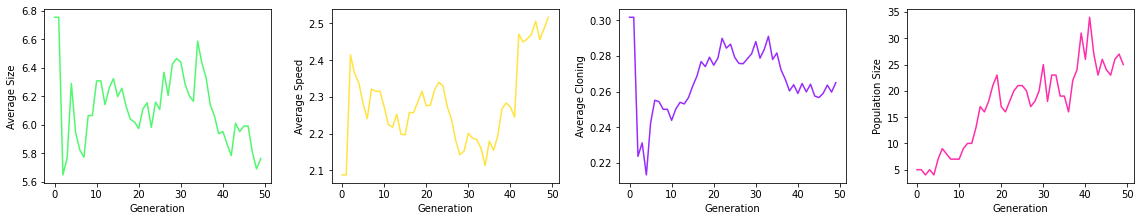}
    \caption{Population Average Speed, Size, Cloning Probability, and Population Growth With 200 Starting Food: Trial 1}
    \label{fig:200Food1}
\end{figure*}
\begin{figure*}[!htb]
    \centering
    \includegraphics[scale=.4]{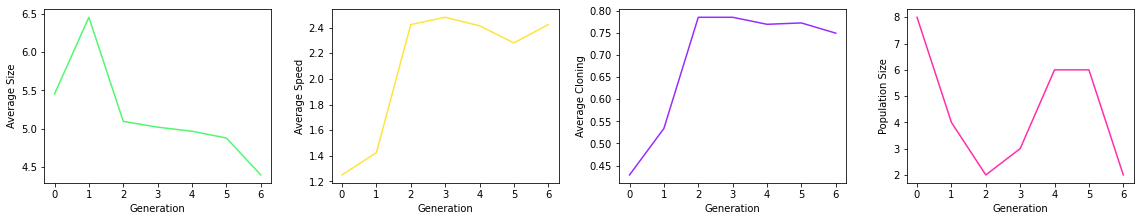}
    \caption{Population Average Speed, Size, Cloning Probability, and Population Growth With 200 Starting Food: Trial 2}
    \label{fig:200Food1}
\end{figure*}\begin{figure*}[!htb]
    \centering
    \includegraphics[scale=.4]{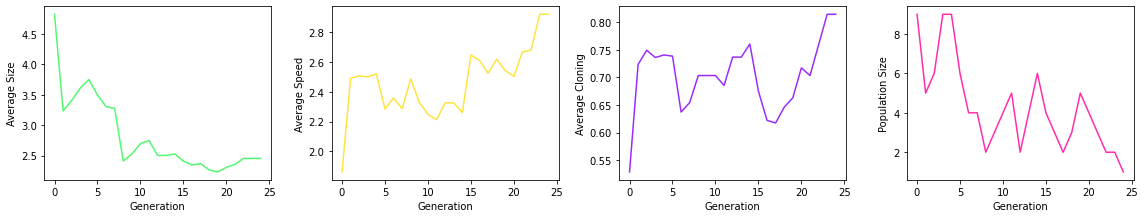}
    \caption{Population Average Speed, Size, Cloning Probability, and Population Growth With 200 Starting Food: Trial 3}
    \label{fig:200Food1}
\end{figure*}
\begin{figure*}[!htb]
    \centering
    \includegraphics[scale=.4]{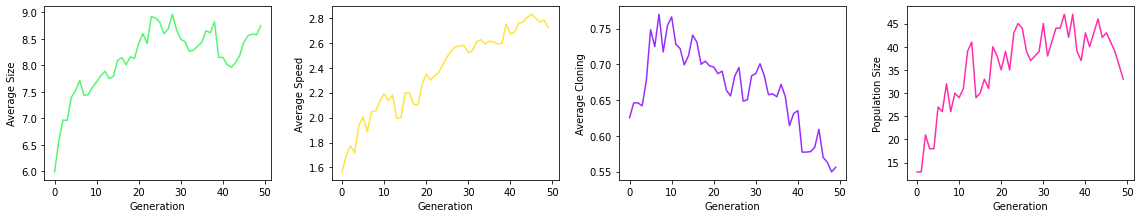}
    \caption{Population Average Speed, Size, Cloning Probability, and Population Growth With 300 Starting Food: Trial 1}
    \label{fig:200Food1}
\end{figure*}
\begin{figure*}[!htb]
    \centering
    \includegraphics[scale=.4]{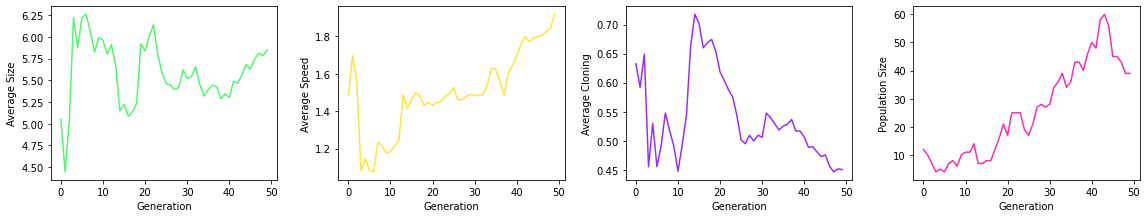}
    \caption{Population Average Speed, Size, Cloning Probability, and Population Growth With 300 Starting Food: Trial 2}
    \label{fig:200Food1}
\end{figure*}
\begin{figure*}[!htb]
    \centering
    \includegraphics[scale=.4]{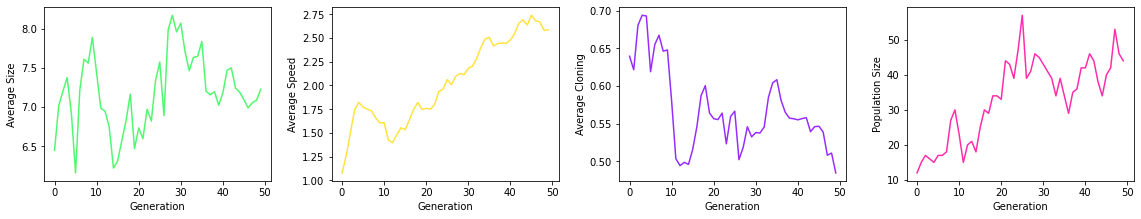}
    \caption{Population Average Speed, Size, Cloning Probability, and Population Growth With 300 Starting Food: Trial 3}
    \label{fig:200Food1}
\end{figure*}

The first observation we made is that when there is a 100 starting food count value in the first run, all of the parameters become constant between the 1st generation and the 15th generation. The second run showed only 4 generations, where all the parameters increased between the first and second generations. This is interesting because that is the same time that the population suddenly decreased, and then increased after the second generation. During the third run, the population seems to have only lived one generation and then all died off after that. The population size seems to either slowly die off, as shown in the first run, or all the entities suddenly die, just like in the second run. The sizes for each run spiked quickly, which tells us that size is an important factor in collecting more food to survive, but even with a bigger size, there should be enough food for the whole population to live. 100 food to start may be too harsh of an environment for most random starting populations to survive. When there was 200 food to start, the population increased dramatically in the first run but not as much in the second and third run. The population in run 2 only survived 6 generations, while the third run lived for 25 generations, which shows us that 200 starting food is not enough to keep the population alive every time. Interestingly, the data for the 200 starting food count and the 300 starting food count seem to be opposite to one another. In the 200 starting food count graph, the population seemed to go up almost consistently, but on the other hand, the 300 starting food count population went up exponentially, and then suddenly decreased between the 40th and the 50th generation. The average size of all the entities never exceeded 6.8, and mostly stayed around 6.2 when there was 200 food, but the average size went all the way up to 9.0 when there was 300 food. The speed when there are 200 starting food went up, then decreased between the 5th and 35th generations, and finally escalated up to 2.5. In contrast, the speed when there is 300 food constantly escalated to about 2.8. Finally, the cloning probability for 200 starting food declined, and at the 5th generation, it rose up to .28, but then declined to .26 in the 40th generation. This trend makes sense because the speed was average between the 10th and 40th generations, so the amount of food each entity collects is consistent with how much they clone, and the population during that time. For 300 food, the cloning probability rapidly increased until about halfway through, and then became mostly constant for the rest of the simulation. All three runs of this lived for all 50 generations, and did not decline at the end. The average speeds for all the entities seems constant compared to the other simulations, and it is obvious that higher speeds are preferred. Still, the cloning probability in every run decreases to below .50, which could mean that the species is probably going to go extinct in the coming generations. An environment with 100 starting food was too harsh on the species and did not give them a good chance to survive. Starting with 200 food, gives us an unstable species where the lifetime is very unpredictable. Therefore, 300 starting food must be the best amount of starting food since the entities consistently lived for all 50 generations, and they have a good population growth.

\section{Key Contributions} \label{contributions}
We created a tool that makes observing evolution and natural selection more efficient so that we can study natural interactions and protect species from extinction. This program also allows us to study viruses and bacterial infections to find ways to make the human body uninhabitable. 

\section{Future Research Directions} \label{results}
In the future, we can code the entities to sexually reproduce, and therefore presenting the possibility for crossover. This will widen the range of mutations and allow us to further investigate evolution and natural selection for a different type of reproduction. We can also use more simulation mechanics, as well as use different types of interactions between entities and their environment. Including more than one type of species can show interactions between the two, which species will outlive the other, and possible crossbreeding if they are biologically close enough. By adding these into our code, we can also include carrying capacity in our environment by including more parameters such as oxygen, space, and water to look at the maximum population that can exist in that environment, Improving our simulations and running more tests will give us more information on each entity and more data on what kind of environment is the best fit for a type of species.

\section{Conclusion}
Using genetic algorithms, we created a tool which can be used for observing natural selection and evolution. By putting a population in an environment with varied amounts of starting food, we were able to collect data on how the species evolved in each environment, and which amount of food led to the most successful batch of entities. We found that the 300 starting food population was the most consistent and did not drastic changes in size, speed, and population size. This tool allows us to add more parameters and different types of species into the environment so we can see how they interact and how each entity grows and evolves over time. Learning about evolution and natural selection is important so that we can learn about how animals live in their natural environment and ways that we can save them from extinction. We can also learn about diseases and how they spread and mutate throughout the body to find ways to make them less of a threat to our health. Using genetic algorithms to simulate natural environments is not only the most efficient way to observe these trends, but is also the safest way, as to not harm people or animals in the process. 

\bibliographystyle{IEEEtran}
\bibliography{bibliography}
\nocite{*}

\end{document}